\documentclass[journal,twoside,web]{ieeecolor}
\usepackage{tmi}
\usepackage{cite}
\usepackage{amsmath,amssymb,amsfonts}
\usepackage{algorithmic}
\usepackage{graphicx}
\usepackage{textcomp}
\usepackage{multirow,booktabs,subcaption}
\usepackage{comment}
\def\BibTeX{{\rm B\kern-.05em{\sc i\kern-.025em b}\kern-.08em
    T\kern-.1667em\lower.7ex\hbox{E}\kern-.125emX}}
\newlength\savewidth\newcommand\shline{\noalign{\global\savewidth\arrayrulewidth
  \global\arrayrulewidth 1pt}\hline\noalign{\global\arrayrulewidth\savewidth}}

\begin{document}

\title{Self-distilled Masked Attention guided masked image modeling with noise Regularized Teacher (SMART) for medical image analysis}
\author{Jue Jiang,  Aneesh Rangnekar, Chloe Min Seo Choi, Harini Veeraraghavan
\thanks{The manuscript has been submitted on XX XX,2024 for review. This work was supported by the MSK Cancer Center core grant P30 CA008748 and NCI R01CA258821.}}

\maketitle

\begin{abstract}
Pretraining vision transformers (ViT) with attention guided masked image modeling (MIM) has shown to increase downstream accuracy for natural image analysis. Hierarchical shifted window (Swin) transformer, often used in medical image analysis cannot use attention guided masking as it lacks an explicit [CLS] token, needed for computing attention maps for selective masking. We thus enhanced Swin with semantic class attention. We developed a co-distilled Swin transformer that combines a noisy momentum updated teacher to guide selective masking for MIM. Our approach called \textsc{s}e\textsc{m}antic \textsc{a}ttention guided co-distillation with noisy teacher \textsc{r}egularized Swin \textsc{T}rans\textsc{F}ormer (SMARTFormer) was applied for analyzing 3D computed tomography datasets with lung nodules and malignant lung cancers (LC). We also analyzed the impact of semantic attention and noisy teacher on pretraining and downstream accuracy. SMARTFormer classified lesions (malignant from benign) with a high accuracy of 0.895 of 1000 nodules, predicted LC treatment response with accuracy of 0.74, and achieved high accuracies even in limited data regimes. Pretraining with semantic attention and noisy teacher improved ability to distinguish semantically meaningful structures such as organs in a unsupervised clustering task and localize abnormal structures like tumors. Code, models will be made available through GitHub upon paper acceptance.
\end{abstract}


\section{Introduction}
\label{sec:intro}

\begin{figure}[!h]
    \centering
    \includegraphics[width=0.8\columnwidth]{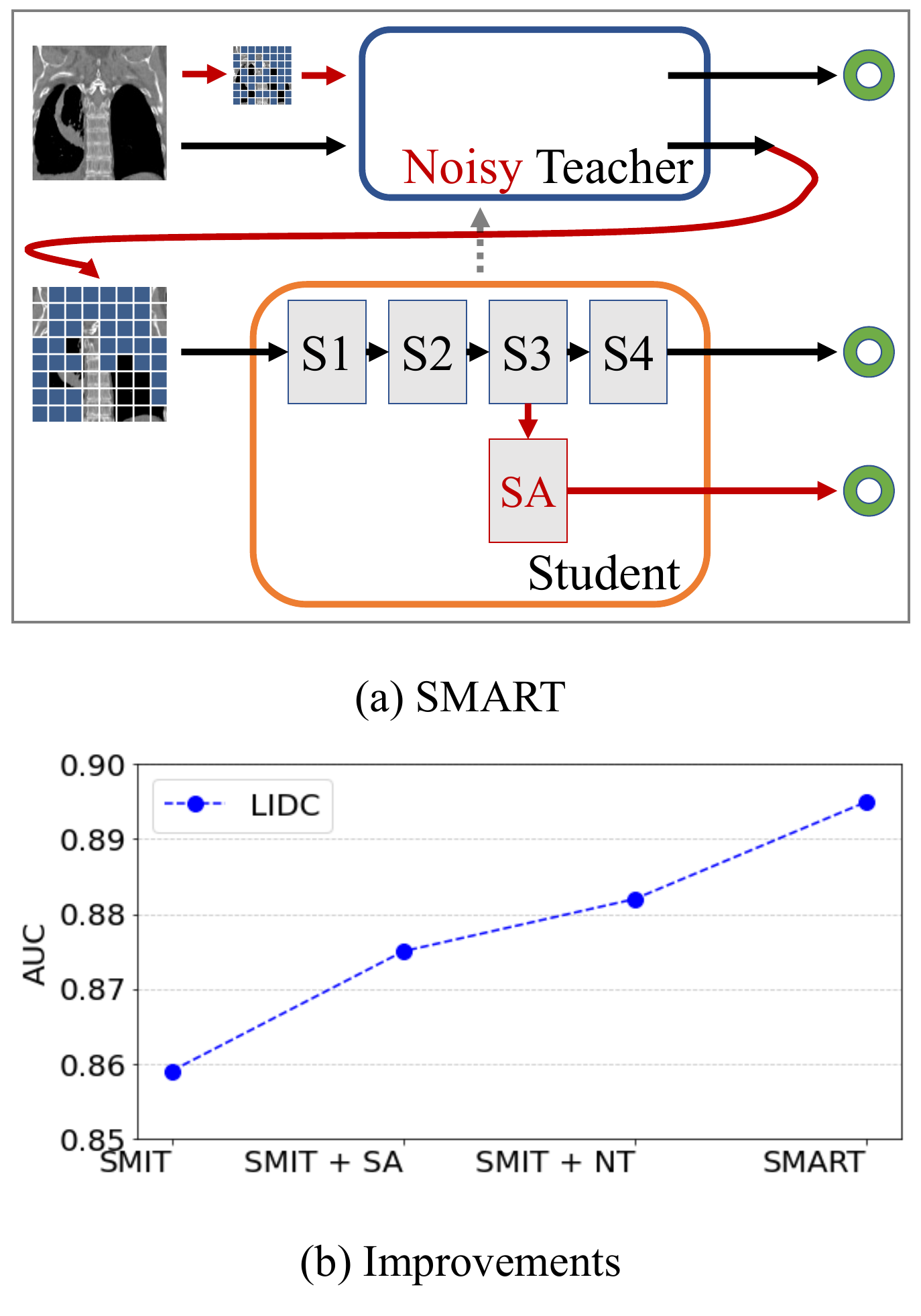}
    \caption{The key components of SMART (a) and the performance on LIDC data with Semantic Attention(SA) and Noisy teacher(NT)}
    \label{fig:figteaserfirstpage}
\end{figure}

Masked image modeling (MIM) with vision transformers (ViT) uses masked patch tokens generated from the input as supervisory signal to guide the extraction of underlying visual semantics within images. This method has shown to be an effective self-supervised learning (SSL) approach for a variety of natural and medical image tasks~\cite{bao2021beit,zhou2022image,jiang2022self_SMIT,li2021mst,liu2021swin}. MIM improves transferability to downstream tasks by increasing diversity of attention heads while maintaining locality inductive bias~\cite{XieCVPR2023_SecretsMIM}. However, the choice of masking strategy crucially impacts accuracy. Attention guided masking in vision transformers has shown to improve the pretraining efficiency and downstream accuracy in natural image analysis applications~\cite{shi2022adversarial_mask,li2022semmae,kakogeorgiou2022hide_attmask,LiuAAAI2023_AttentionThrow}. 

In medical image applications, MIM-based self-supervised pretraining of hierarchical shifted window (Swin~\cite{liu2021swin}) transformers has been shown to produce more accurate segmentation (or dense pixel predictions) than vision transformers and convolutional networks~\cite{tang2022self,jiang2022self_SMIT}. However, such attention guided masking pretext tasks are lacking for the Swin~\cite{liu2021swin} transformer architecture, mainly due to an absence of the global class token (denoted as [CLS]), which is essential for aggregating attention across the diverse visual elements present in an image.

Hence, we enhanced the Swin architecture for attention guided MIM (Fig.~\ref{fig:figteaserfirstpage}). Concretely, we introduce a semantic attention module after stage \#3 of Swin~\cite{liu2021swin} by adding a ViT self-attention block with the global semantic token [CLS]. We use global self-attention to enable Swin to extract global image-level attention for selective masking. Following AttMask~\cite{kakogeorgiou2022hide_attmask}, our approach masks high-attending patch tokens in order to force the student to learn the embedding of salient parts of images. 

We adopted co-distillation with MIM~\cite{caron2021emerging} that combines a dynamically updated momentum teacher with identical architecture as the student network. Co-distillation obviates need for a pretrained high-capacity teacher~\cite{pmlr-v139-touvron21a,XueCVPR_2023} and has been shown effective for attentive masking with ViT~\cite{kakogeorgiou2022hide_attmask}. However, as the momentum teacher performs the selective masking for student, there is a possibility that it masks tokens that makes token distillation simple, leading to a potential leak between the student and the teacher. Hence, we introduced the noisy teacher scheme to increase the diversity of tokens and thereby, better self-supervised pretraining. We hypothesized that attention guided MIM, combined with noisy teacher regularization, would improve performance on downstream tasks. 

Our contributions are: (a) an approach for global attention extraction for Swin using semantic attention which allows attention guided masking in a co-distillation framework, (b) semantic attention for directly calculating attention maps to perform and visualize downstream task with Swin, (c) a noisy teacher regularizer in distillation framework, (d) Finally, we performed comprehensive analysis of the impact of attention guided MIM and noisy teacher on many and data limited regimes, capability of pretrained models to distinguish semantically meaningful structures in images as well as segment tumors in zero-shot manner. 

\section{Related works}
\subsection{Masking strategies for MIM}
Random masking~\cite{he2021masked,jiang2022self_SMIT,xie2021simmim,zhou2022image} is a commonly used masked image modeling approach for pretraining ViT~\cite{dosovitskiy2021} and Swin~\cite{liu2021swin} networks. Although computationally simple to implement, random masking of image patches can leave highly correlated patches including spatially adjacent patches visible, which may reduce it's effectiveness. Blockwise matching~\cite{wei2022masked_feature, bao2021beit, wang2022bevt} mitigates this issue by grouping blocks of spatially adjacent patches while using a computationally simple approach without requiring additional networks to select regions for masking. However, the masked regions are still selected randomly, which can disperse attention towards irrelevant portions of the image, thus wasting computational resources used for pretraining.
\\
Attentive masking methods alleviate the aforementioned difficulty by focusing the pretraining task towards relevant image portions, which increases accuracy. Selective masking is often accomplished by aggregating attention from the [CLS] token used to predict image-level class in the ViT~\cite{li2021mst,kakogeorgiou2022hide_attmask,LiuAAAI2023_AttentionThrow}. This approach requires no additional pretrained networks as is needed in semantic attention methods~\cite{li2022semmae}. Thus, we modify the Swin transformer to enable global self-attention and thereby, perform attentive masking.

\subsection{Noise injected distillation learning}
Noise injection, commonly implemented via dropout~\cite{liu2023patchdropout,lee2023self,tarvainen2017mean} is used to diversify the extracted tokens as well as reduce spatial locality of extracted features. Noise injection using dropout is used to align predictions from different token extracted by the same network in self-distillation~\cite{lee2023self} as well as align features in knowledge distillation frameworks~\cite{arani_FickleTeacherWACV2021,liu2020noisy,bulo2016dropout}. Noise injection has also shown to increase accuracy of self-supervised learning methods for classification~\cite{tarvainen2017mean, laine2016temporal} and dense prediction tasks~\cite{you2022simcvd}. 
\\
Noise-based regularization is effective because it provides a form of implicit data augmentation~\cite{you2022simcvd} and improves training convergence and generalization on unseen testing data by allowing the network to escape from local minima~\cite{zhou_PMLR_2019}. In addition, noise regularized knowledge distillation has shown to improve both in- and out-of-distribution accuracy~\cite{arani_FickleTeacherWACV2021}. Finally, noise injected into plain ViT through patch dropout~\cite{liu2023patchdropout} was shown to reduce memory requirements in training while improving training data utilization.
\\

\begin{figure*}[t]
    \centering
    \includegraphics[width=0.9\textwidth]{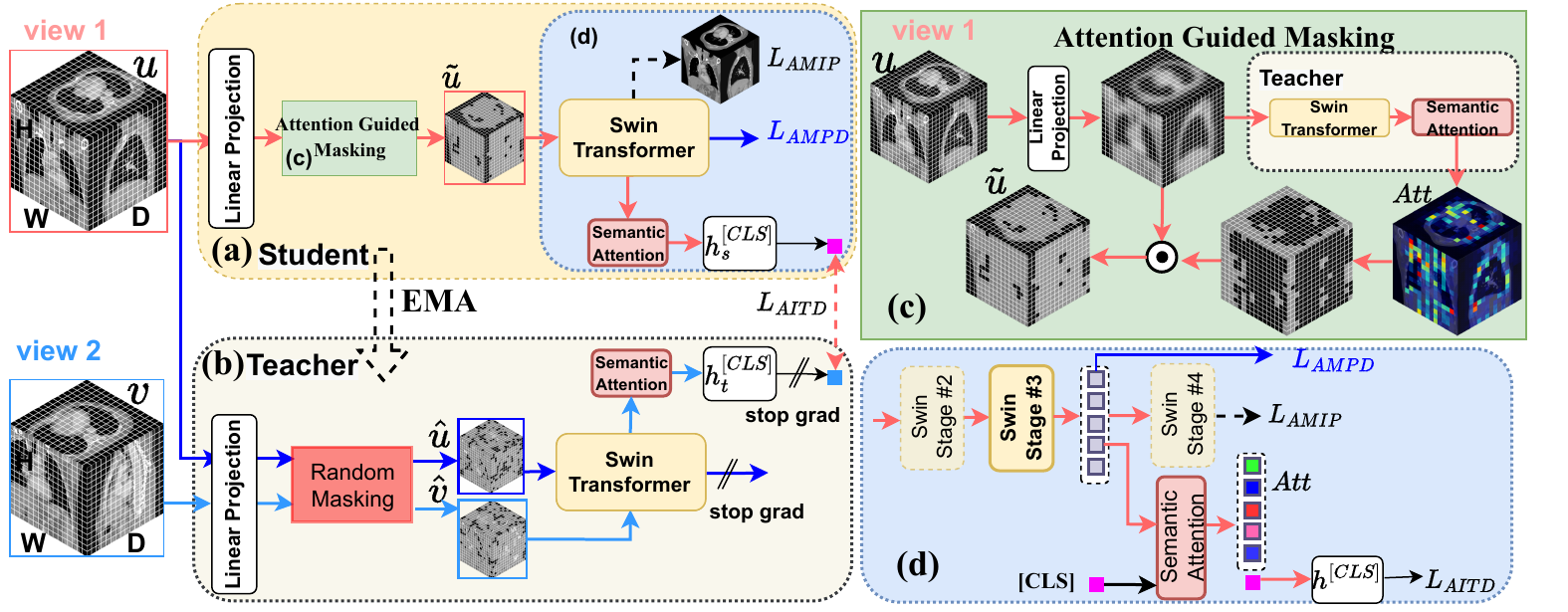}
    \caption{SMART: (a) shows the attention-guided masking for student and (b) shows the noisy teacher for the exponential moving average(EMA) process. (c) shows the detailed attention guided masking for student and   (d) depicts the different outputs produced by newly-modified Swin transformer. We use [CLS] to represent the class token embedding obtained by integrating the Semantic Attention (SA) layer into stage \#3 of the Swin transformer. We use an additional prediction head (Decoder) for the masked image prediction.}
    \label{fig:method}
\end{figure*}

\section{Method}

\subsection{Overview} We present our pretext self-supervised learning approach for Swin transformers: \textbf{S}e\textbf{M}antic \textbf{A}ttention guided co-distillation with \textbf{R}egularized noisy \textbf{T}eacher (SMART), which builds and improves a Swin transformer self-supervision approach called SMIT~\cite{jiang2022self_SMIT}. Similar to SMIT, SMART uses a co-distillation framework consisting of a student ($f_{s}(\theta_{s})$) and a teacher ($f_{t}(\theta_{t})$) network with identical model architecture (Fig~\ref{fig:method}). $f_{t}(\theta_{t})$ is updated via exponential moving average of $f_{s}(\theta_{s})$ and following SSL, $f_{s}(\theta_{s})$ is adopted and fine-tuned for downstream tasks.

\subsection{Semantic attention guided masked image modeling}
The Swin transformer architecture performs self-attention within defined spatial windows, with shifted windows partioning and cross window relationships. Hence, Swin is fundamentally not designed to extract global attention between all the patches, which is required to perform attention guided masking. We remedy this issue by inserting a semantic attention module, integrated with a gloabl image semantics token (denoted as [CLS]), as illustrated in Fig.~\ref{fig:method}a. 
\\
The SA module is composed of a 2-layer ViT block consisting of a multi-head attention layer followed by a feed forward attention network with the \textit{LayerScale \/}\rm~\cite{touvron2021CAIT}. SA combines the inserted [CLS] with $N$ patch tokens to create a tokenized feature z $\in$ $\mathbb{R} ^{(N+1)\times D}$, with $\{z_{i}\}_{i=1}^{N}$ being embedding for patch ${i}$, and $z_{[CLS]}=z_{N+1}$. $D$ is the feature size. The multi-head self attention with $h$ heads and $N+1$ patches and [CLS] as query, uses the projection matrices $W_{q}$, $W_{k}$, $W_{v}$ to compute the query $Q^{[CLS]} = W_{q} \times z_{[CLS]} + b_{q}$, key $K = W_{k} \times z + b_{k}$, and value $V = W_{v} \times z + b_{v}$ features, where $b_q, b_k, b_v$ are the biases. Next, the semantic attention (SATT) is computed by averaging over all the attention heads:
\begin{equation}
\setlength{\abovedisplayskip}{1pt}
\setlength{\belowdisplayskip}{1pt} 
SATT = \frac{1}{h} \sum_{1}^{h} \rm {Softmax}(Q_h^{[CLS]} \cdot \frac{K_h^T}{\sqrt{D/h}}). 
\label{eqn:attmsk}
\end{equation}  
$SATT$ $\in$ [0,1] is a vector of size N, corresponding to the patch indices, and where larger values indicate higher attention. $SATT$ can thus be used to mask elements within the mask vector $m^{satt}$ to produce attention guided masked tokens $\tilde{u} = m^{satt} \odot u$ that is then input to the student network $f_s(\theta_{s})$ (Fig.~\ref{fig:method}c).
\\
Selective masking allows to implement both low attending~\cite{li2021mst} and high attending~\cite{kakogeorgiou2022hide_attmask} approaches previously developed with ViT models. The SMART approach for Swin masks high attending tokens by setting the indices of high attending tokens within $m^{att}$ to 0. Thus, SATT is sorted in descending order to identify the top [$rN$], where $r \in [0,1]$ is the masking ratio. We also keep a small number $s$ (with $s < r$) of the top high-attending or hint tokens unmasked to provide additional information for the masked prediction pretext task. We used $r$ of 0.7 and $s$ of 0.1 in all our experiments. The masked input for $f_{s}(\theta_{s})$ is thus generated as $\tilde{u} = m^{satt} \odot u$.  

\subsection{Co-distillation framework}
Two different augmented views $u$ and $v$, produced by random cropping of 3D regions from input CT volume are converted into a sequence of of patch tokens $\{u_{i}\}_{i=1}^{N}, \{v_{i}\}_{i=1}^{N}$, $N$ being the number of image patches, masked (random for $f_{t}(\theta_{t})$ and attentive for $f_{s}(\theta_{s})$) and then input to the networks (Fig.~\ref{fig:method}). The networks parameters are updated using co-distillation losses consisting of attention guided masked patch token distillation (AMPD), attention guided image token distillation (AITD), global image token distillation (GITD). In addition, the attention guided masked image prediction (AMIP) loss is used only for student model. 
\\
\noindent\textbf{Attention guided image token distillation (AITD) \/}\rm is unique to SMART. It measures the dissimilarity in the token distributions $P_{s}^{[CLS]}$ and $P_{t}^{[CLS]}$ produced by the SA blocks (Fig.~\ref{fig:method}a) in teacher and student networks. $P_{s}^{[CLS]}$ and $P_{t}^{[CLS]}$ are produced by linear projection layers $h_s^{[CLS]}(f_{s}(\theta_{s},\tilde{u}))$ and $h_{t}^{[CLS]}(f_{t}(\theta_{t}, \hat{v}))$ using masked tokens $\{ \tilde{u}, \hat{v} \}$. AITD is defined as:
\begin{equation}
\setlength{\abovedisplayskip}{1pt}
\setlength{\belowdisplayskip}{1pt} 
\begin{split}
L_{AITD} = - \sum_{i=1}^N P_{t}^{[CLS]}(\hat{v}_{i},\theta_t) log (P_{s}^{[CLS]}(\tilde{u}_{i},\theta_s)).
\end{split}
\end{equation}
The sharpness of the token distribution is controlled using a separate temperature term $\tau_{s} >0$ and $\tau_{t} > 0$ for the student and teacher networks, respectively. Sharpening transform is expressed using student network as:
\begin{equation}
\setlength{\abovedisplayskip}{1pt}
\setlength{\belowdisplayskip}{1pt} 
    P_{s}^{[CLS]}(\tilde{u},\theta_{s}) = \frac{exp(h_s^{[CLS]}(f_{s}(\tilde{u}_{j},\theta_{s}))/\tau_{s}}{\sum_{j=1}^{K}exp(h_s^{[CLS]}(f_{s}(\tilde{u}_{j},\theta_{s}))/\tau_{s}}.
\end{equation}
\\
\noindent\textbf{Global image token distillation (GITD)  \/}\rm measures similarity of the global image token distributions $P_{s}^{[g]}$ and $P_{t}^{[g]}$ produced from the average pooling layer placed after stage \#4 and generated from two corrupted views $\tilde{u}$ and $\hat{v}$, for student and teacher, respectively. Unlike SMIT~\cite{jiang2022self_SMIT}, GITD loss is computed using masked tokens for both teacher and student as:
\begin{equation}
\setlength{\abovedisplayskip}{1pt}
\setlength{\belowdisplayskip}{1pt} 
\begin{split}
L_{GITD} = - \sum_{i=1}^N P_{t}^{[g]}(\hat{v}_{i},\theta_t) log (P_{s}^{[g]}(\tilde{u}_{i},\theta_s)).
\end{split}
\end{equation}
Similar to AITD loss, the token distributions are subjected to sharpening transforms. The key difference of GITD and AITD is that the former is computed after \#S4 layer while the latter is computed using the SA layer output (Fig.~\ref{fig:method}c). 
\\
\noindent\textbf{Attention guided masked patch prediction (AMPD) \/}\rm loss measures the dissimilarity in the masked patch token distributions $P_t^{Patch}$ and $P_s^{Patch}$ produced by $f_{t}(\hat{u}, \theta_t)$ and $f_{s}(\tilde{u}, \theta_{s})$ using stage \#3 output of the Swin before the semantic attention layer (Fig.~\ref{fig:method}c): 
\begin{equation}
\setlength{\abovedisplayskip}{1pt}
\setlength{\belowdisplayskip}{1pt} 
\begin{split}
L_{AMPD} = - \sum_{i=1}^N m_i \cdot  P_{t}^{Patch}(\hat{u}_{i},\theta_t) log (P_{s}^{Patch}(\tilde{u}_{i},\theta_s)),
\end{split}
\end{equation}
where, $P_s^{Patch}$ and $P_t^{Patch}$ are computed by applying \it{softmax \/}\rm to the outputs of $h_s^{Patch}$ and $h_t^{Patch}$ and before performing patch merging of stage \#3 output. Sharpening transforms are also applied to the token distributions. Of note, this loss measures the similarity of the generated tokens for the same view corrupted in different ways before presenting to student and teacher. 
\\
\noindent\textbf{Attention guided masked image prediction (AMIP) \/}\rm measures the reconstruction error in the image generated by $f_{s}(\theta_{s})$. The reconstructed view $u^{\prime}$ is produced by projecting the masked input $\tilde{u}$ through a 1-layer linear projection layer $h_{s}^{Pred}$. The loss is computed as: 
\begin{equation}
\setlength{\abovedisplayskip}{1pt}
\setlength{\belowdisplayskip}{1pt} 
L_{AMIP} = \sum_{i}^{N} E\| m_{i}^{att} \cdot (h_{s}^{Pred}(f_{s}(\tilde{u_{i}}, \theta_{s})) - u^{\prime}_{i}) \|_{1}, 
\end{equation} 
where, $m_{i}^{att}$ is the masked token vector produced through attention guided masking using the semantic attention layer.  The total loss is computed as, $L_{total}$ = $L_{AMIP}$ + $\lambda_{AMPD}$ $L_{AMPD}$ + $\lambda_{AITD}$ $L_{AITD} + \lambda_{GITD} G_{ITD}$. 

\subsection{Noisy teacher}
\label{subsec:noisyteacher}
Noisy teacher was implemented by using patch dropout~\cite{liu2023patchdropout}, which randomly masks image patch tokens presented to the teacher. Concretely, elements within mask vectors $m_u, m_v$ corresponding to indices of patch tokens for $u$ and $v$ are randomly set to 0, using a patch drop ratio $r_t \in [0,1]$, where $r_t = 0.7$. The corrupted inputs $\hat{u} = m_u \odot u$ and $\hat{v} = m_v \odot v$ are thus generated for $f_{t}(\theta_{t})$. The token embeddings are generated by the teacher network followed by \textit{softMax \/}\rm activation prior to applying attention guided masking to the student network. The teacher network was updated using exponential moving average as $\theta_{t}=\lambda_m\theta_{t} + (1-\lambda_m)\theta_{s}$, where $\lambda_{m}$ is momentum, and a cosine schedule from 0.996 to 1 (Fig.~\ref{fig:method}b).

\section{Experiments and Results}
Experimental comparisons were performed against SMIT~\cite{jiang2022self_SMIT}, which uses random masking with a Swin backbone as well as ViT-based AttMask~\cite{kakogeorgiou2022hide_attmask} to assess accuracy gains with semantic class attention and noisy teacher regularization used in SMART. We also compared against other masking strategies including blockwise masking using iBot~\cite{zhou2022image} and low attending masking using masked self-supervised transformer (MST~\cite{li2021mst}) on ViT~\cite{dosovitskiy2021}.

\subsection{Pretraining datasets}
\label{subsec:pretrainingdatasets}
All models were pretrained on a diverse set of 10,412 3D CT scans. These scans encompass a range of diseases, including COVID-19, as well as abdominal, lung, head and neck, kidney, and liver cancers. The data for these scans were sourced from both institutional and public repositories, and no segmentations were used during the self-supervised pretraining phase.
\\
The individual dataset descriptions are: 
\begin{itemize}
    \item \textbf{MSKCC Lung Cancer Image CT} includes 5,124 thoracic CT scans, collected internally from patients with non-small cell lung cancers undergoing image guided radiation treatment.
    \item \textbf{MSKCC Head and Neck CT} comprises 2,632 internal head and neck CT image scans from patients diagnosed with cancers in one of nasopharynx, oropharynx, or hypopharynx and who underwent image guided radiation treatment.
    \item \textbf{MICCAI 2022 MELA Challenge} includes 880 image scans from MICCAI 2022 MELA Challenge~\cite{xiao2023lesion} for the lung Mediastinal Lesion Analysis.
    \item \textbf{AMOS 2022 Challenge} contains 360 scans from the AMOS22 dataset~\cite{ji2022amos}, designed to facilitate abdominal multi-organ segmentation.
    \item \textbf{COVID-19 CT} contains 609 CT scans, including patients diagnosed with COVID-19 and control patients without COVID-19. This data was collected from multiple institutions in the United States and China and is publicly available~\cite{harmon2020artificial}.
    \item \textbf{TCIA NSCLC} includes 316 image scans of patients with non-small cell lung cancer (NSCLC) patients, obtained from~\cite{aerts2015data}.
    \item \textbf{Kidney Cancer CT} comprises of 411 image scans from the Cancer Genome Atlas Kidney Renal Clear Cell Carcinoma Collection (TCGA-KIRC~\cite{akin2016radiology}). 
    \item \textbf{Pancreas CT}  consists of 80 contrast-enhanced abdominal CT volume scans focusing on pancreas segmentation, described in~\cite{roth2015deeporgan}.
\end{itemize}

\subsection{Downstream datasets and tasks}
\label{subsec:finetuningdatasets}

We evaluated the following downstream tasks for studying the benefits of using SMART pretraining strategy. \\
\noindent\textbf{Task 1: Distinguish 3D organs using pretrained features}
We evaluated our pretrained model's capability to differentiate various organs on the OrganMNIST3D dataset, which consists of 1,743 3D CT scans representing 11 different organs in the abdomen.
\\
\noindent\textbf{Task 2: Binary classification of Malignant and Benign lung nodules from patients screened for lung cancer (LC)}

The Lung Image Database Consortium (LIDC) dataset~\cite{armato2011lung} consists of 1,010 patients from seven different institutions, with a total of 2,426 lung nodules extracted using pylidc library. Each nodule was rated on a malignancy from scale 1-5, with ratings of 1$-$3 as benign and 4$-$5 as malignant, following the approach of~\cite{chen2019med3d}. This grouping approach resulted in 2,054 benign and 540 malignant nodules. We employed a 3-fold stratified cross validation on 1624 nodules, and used the rest 1,000 as test set.
\\
\noindent\textbf{Task 3: Binary prediction of immunotherapy response}
We defined response $\geq$ 6 months after treatment completion vs non durable benefit [NDB]) to LC treatment from pre-treatment CTs. The dataset, gathered at MSKCC, consists of two hundred Stage IV patients diagnosed with non-small cell lung cancer, who were treated with standard of care contrast-enhanced CT of the chest performed pre-treatment and containing at least one expert-delineated visible tumor in the lung. The response distribution of Durable Clinical Benefit [DCB] to Non-Durable Benefit [NDB] is 82:118. Three-fold stratified cross validation was applied.
\\
\noindent\textbf{Task 4: Zero-shot Disease Localization}\\
We used the TCIA 5rater dataset~\cite{Wee2019} which consists of 20 LC patients treated with image-guided radiotherapy (IMRT) and segmented by five radiation oncologists (5R).
\\

\subsection{Implementation details}
\label{sup_subsec:implementationdetails}

MONAI~\cite{cardoso2022monai} and Pytorch~\cite{Paszke2019PyTorchAI} libraries were used for implementation and training of the various models. \textbf{For self-supervised learning,} SMART and various comparable methods such as AttMask~\cite{kakogeorgiou2022hide_attmask} (\includegraphics[height=0.7em]{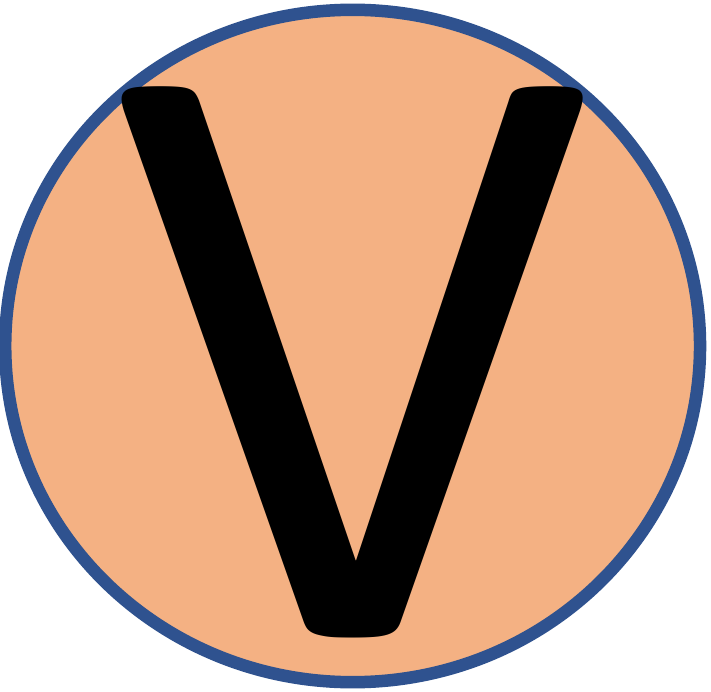}), MST~\cite{li2021mst} (\includegraphics[height=0.7em]{emojis/logo_vit.pdf}), iBot~\cite{zhou2022image} (\includegraphics[height=0.7em]{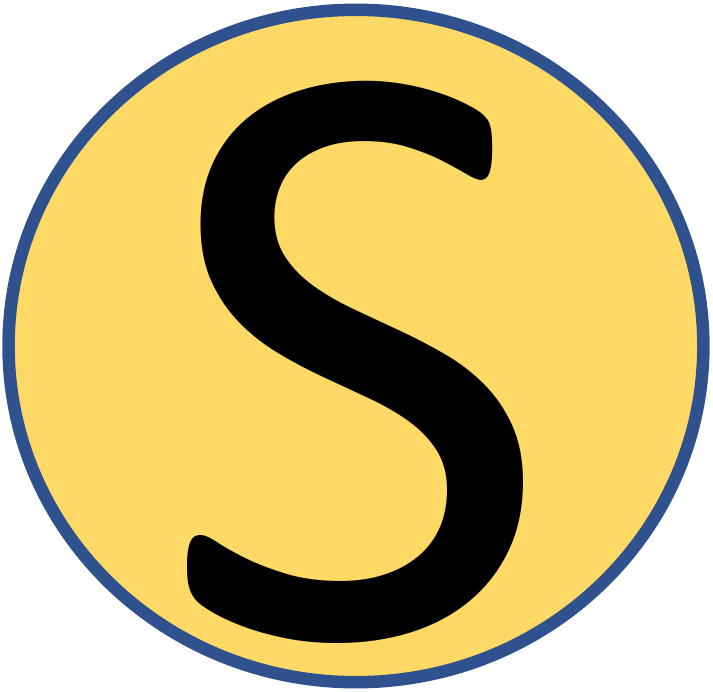}) and SMIT~\cite{jiang2022self_SMIT} (\includegraphics[height=0.7em]{emojis/logo_swin.pdf}) were pretrained on datasets as mentioned in Section \ref{subsec:pretrainingdatasets}.

Two different transformer structure were evaluated in this study, ViT and Swin. The ViT~\cite{dosovitskiy2021} architecture  (\includegraphics[height=0.7em]{emojis/logo_vit.pdf}) comprised of 12 transformer blocks, 768 embedding features, and 8 multi-head self attention. The Swin~\cite{liu2021swin} architecture (\includegraphics[height=0.7em]{emojis/logo_swin.pdf}) used a depth of [2,2,8,2] and [4,4,8,16] multi-head for each transformer depth, and a feature embedding size of 384. This setup also included a window size of 4 $\times$ 4 $\times$ 4 and patch size of 2 $\times$ 2 $\times$ 2.
 
We generated augmented views by re-sampling the scans at 2mm $\times$ 2mm $\times$ 2mm voxel spacing and then randomly cropping 128 $\times$ 128 $\times$ 128 voxel scans. The networks were optimized using ADAMw~\cite{Loshchilov2017DecoupledWD} with a cosine learning rate scheduler~\cite{Loshchilov2016SGDRSG} trained for 800 epochs with an initial learning rate of $8e^{-4}$ and warmup for 80 epochs. A path drop rate of 0.1 was applied to the student model, and pretraining was conducted on four NVIDIA A100 GPUs (each with 80GB memory). Parameters $\lambda_{AITD}$ = 0.1, $\lambda_{GITD}$ = 0.1, amd $\lambda_{AMPD}$ = 0.1 in Section \ref{subsec:noisyteacher} were determined experimentally via grid-search. Degenerate solutions were avoided using centering and sharpening operations~\cite{caron2021emerging,jiang2022self_SMIT}. 

\textbf{For fine-tuning,} we used a single NVIDIA A40 GPU throughout all fine-tuning experiments. For LIDC classification (Task 3), we used a batch size of 40 with a learning rate of $2e^{-4}$ for 1,000 epochs. For Immunotherapy classification (Task 4), we used a batch size of 40 with a learning rate of $2e^{-4}$ for 2,000 epochs.

\subsection{Evaluation metrics}
Task 1 was evaluated using inter-cluster and intra-cluster distance, computed using the Euclidean Distance of different classes; Task 1 and 2 are evaluated using The Area Under the Curve (AUC); Task 4 was evaluated using Dice-Sørensen coefficient(DSC).

\section{Results}

\begin{figure*}[htp]
    \centering
    \includegraphics[width=0.7\textwidth]{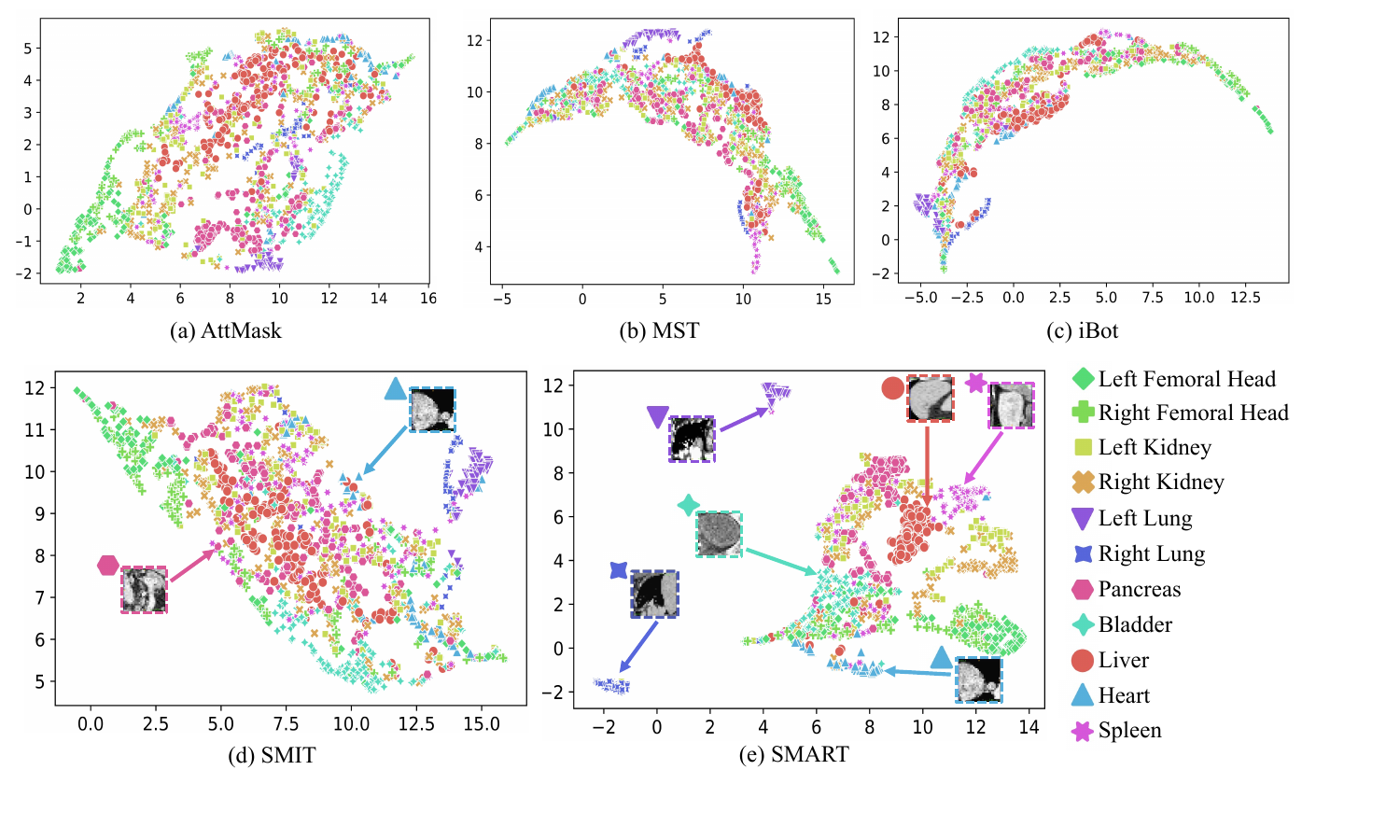}
    \caption{UMAP-based clustering of features extracted by various analyzed pretrained models applied to the OrganMNIST data without additional fine tuning.}
    \label{fig:umap_dist_models}
\end{figure*}

\subsection{Unsupervised clustering to identify distinct organs}
We performed unsupervised clustering of image tokens for SMIT(\includegraphics[height=0.7em]{emojis/logo_swin.pdf}) and [CLS] token for SMART(\includegraphics[height=0.7em]{emojis/logo_swin.pdf}) as well as ViT-based methods to extract salient image regions from medical images without doing any fine tuning. As shown in Table.~\ref{tab:umap_dist_models}, SMART produced the best separation of the different organs resulting in the lowest intra-class and highest inter-class distances. SMIT was second most accurate pretext-task method. The detailed group UMAP distribution of the compared methods are shown in Fig. \ref{fig:umap_dist_models}. As shown, SMART produced distinct clusters for the individual organs. The separation of the clusters for the various organs was worse than SMART for all other methods.

\begin{table}[t]
\def\arraystretch{1.2}
\scriptsize
\centering
\caption{Intra and inter-cluster distances of unsupervised clustering produced for various masking methods on the OrganMNIST dataset. Backbone: ViT (\includegraphics[height=0.75em]{emojis/logo_vit.pdf}) and Swin (\includegraphics[height=0.75em]{emojis/logo_swin.pdf}).}
\label{tab:umap_dist_models}
\resizebox{\columnwidth}{!}{%
\begin{tabular}{lll}
Pretext SSL Task & Intra-cluster ($\downarrow$) & Inter-cluster ($\uparrow$) \\ \shline
AttMask~\cite{kakogeorgiou2022hide_attmask} \includegraphics[height=1em]{emojis/logo_vit.pdf} & {1.59 $\pm$ 0.41} & {2.95 $\pm$ 1.37} \\
MST~\cite{li2021mst} \includegraphics[height=1em]{emojis/logo_vit.pdf}  & {2.22 $\pm$ 0.47} & {3.47 $\pm$ 1.98} \\
iBot~\cite{zhou2022image} \includegraphics[height=1em]{emojis/logo_swin.pdf} & {2.13 $\pm$ 0.37} & {3.29 $\pm$ 2.07}\\
SMIT~\cite{jiang2022self_SMIT} \includegraphics[height=1em]{emojis/logo_swin.pdf} & {1.53 $\pm$ 0.45} & {3.73 $\pm$ 2.74}\\
\hline
\hspace{0.5em}SMART wo NT \includegraphics[height=1em]{emojis/logo_swin.pdf} &1.43$\pm$0.49& 4.18$\pm$2.74 \\
\hspace{0.5em}SMART wo SA \includegraphics[height=1em]{emojis/logo_swin.pdf} &1.46$\pm$0.30 & 5.17$\pm$2.98 \\
\hspace{0.5em}\textbf{SMART} \includegraphics[height=1em]{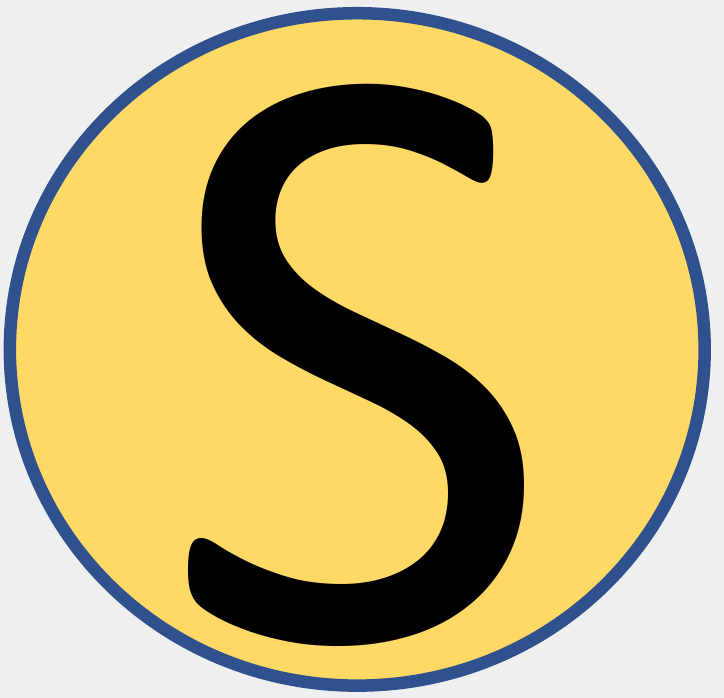} & \textbf{1.28 $\pm$ 0.57} & \textbf{5.93 $\pm$ 3.30}\\ \hline
\end{tabular}%
}
\end{table}

\subsection{SMART are more accurate for lung nodule classification}
\label{sup_sec:Classification_LIDC}
Tables \ref{tab:results_classification_lidc} demonstrate that SMART (\includegraphics[height=0.7em]{emojis/logo_swin.pdf}) surpassed all other methods in lung nodule classification, showing a significant accuracy improvement over both SMIT (\includegraphics[height=0.7em]{emojis/logo_swin.pdf}) and ViT-based AttMask (\includegraphics[height=0.7em]{emojis/logo_vit.pdf}) methods. Overall, Swin-based methods were more accurate than ViT-based methods. Specifically, SMART achieved an AUC of 0.642 with linear probing and an AUC of 0.895 with fine-tuning.

\subsection{SMART are more accurate for treatment reponse prediction}
\label{sup_sec:Classification_IO}

Tables\ref{tab:results_classification_immunotherapy} indicate that SMART (\includegraphics[height=0.7em]{emojis/logo_swin.pdf}) outperformed all other methods in predicting treatment response, showing a noticeable accuracy improvement over both SMIT (\includegraphics[height=0.7em]{emojis/logo_swin.pdf}) and ViT-based AttMask (\includegraphics[height=0.7em]{emojis/logo_vit.pdf}) methods. Generally, Swin-based methods were more accurate than ViT-based ones. Specifically, SMART achieved an AUC of 0.660 with linear probing and an AUC of 0.740 with fine-tuning.

\subsection{Semantic Attention map}
\begin{figure}[t]
    \centering
    \includegraphics[width=\columnwidth]{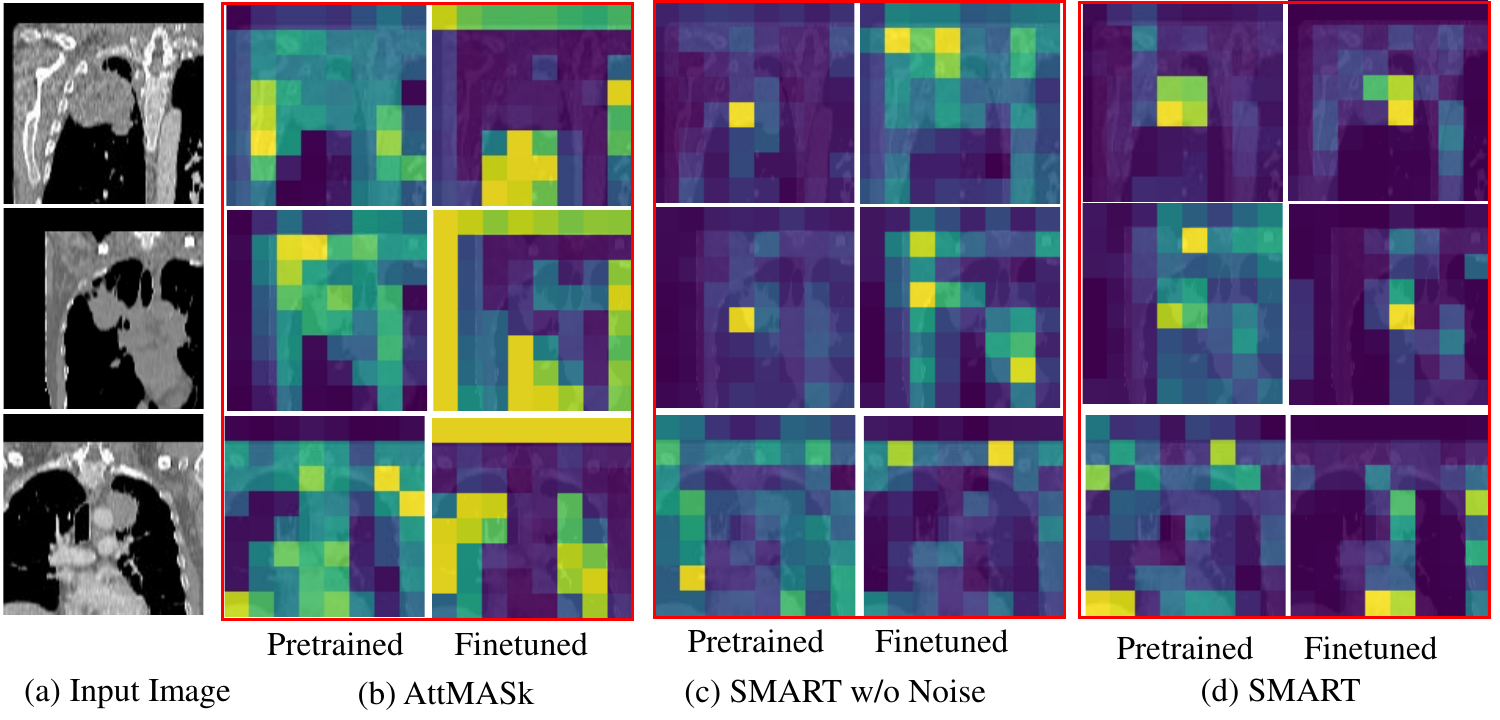}
    \caption{Attention maps computed for example images by AttMask (b,c) and SMART (d,e) after pretraining and fine tuning on the Immunotherapy dataset for response prediction. Tumors are indicated within red bounding boxes. }
    \label{fig:semanticattn_eval_main}
\end{figure}

Fig.~\ref{fig:semanticattn_eval_main} shows example images with attention maps generated using AttMask and SMART as well as attention guided masks produced by the teacher network during pretraining. Random masking inconsistently masked salient regions such as lungs and tumor compared to AttMask and SMART. SMART was more consistent in masking tumors, while AttMask focused on the lungs in the top but around the middle section containing the aorta in the bottom image. 

Fig.~\ref{fig:semanticattn_eval_main} show the attention maps computed on the images after fine-tuning for Immunotherapy response prediction. In all cases, SMART accurately predicted the Immunotherapy outcome as DCB, NDB, and DCB. AttMask generated inaccurate predictions for all three cases, focused mostly on the lungs and heart. SMART attention map visualization clearly show that the high attending regions are on the tumor, which is most relevant for predicting cancer treatment outcomes. 

\begin{figure}[t]
    \centering
    \includegraphics[width=\columnwidth]{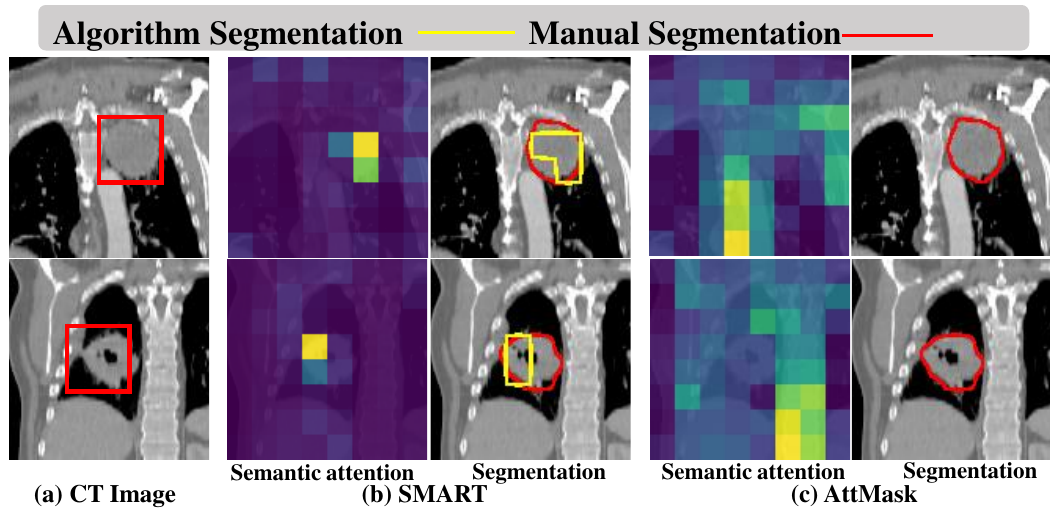}
    \caption{Attention map for AttMask (b) and SMART (c) for zero-shot segmentation of lung tumor (a) on an example from the zero-shot 5R dataset. Tumor is indicated within red bounding box.}
    \label{fig:semanticattn_eval_zeroshot}
\end{figure}

\subsection{Zero-shot Disease Localization}
We further evaluated the ability to zero-shot localize lung cancer (LC) tumors using attention maps from the pretrained models, AttMask(\includegraphics[height=0.7em]{emojis/logo_vit.pdf}) and SMART(\includegraphics[height=0.7em]{emojis/logo_swin.pdf}) on an independent dataset consisting of 21 lung tumors treated with radiotherapy. LC segmentation was produced by thresholding the attention maps using 90\% percentile of the attention values. Our analysis showed that both methods were applicable to detect tumors in zero-shot setting with SMART(\includegraphics[height=0.7em]{emojis/logo_swin.pdf}) generating more accurate LC segmentations with DSC of 0.62 $\pm$ 0.32 than AttMask(\includegraphics[height=0.7em]{emojis/logo_vit.pdf}) with DSC of 0.51 $\pm$ 0.34.

\begin{table}[t]
\caption{LIDC (1,082/542/1,000)}
\label{tab:results_classification_lidc}
\def\arraystretch{1.25}
\resizebox{0.48\textwidth}{!}{%
\begin{tabular}{lllllll}
\multirow{2}{*}{\begin{tabular}[c]{@{}l@{}}Pretext SSL\\ Task\end{tabular}} & \multicolumn{3}{c}{Linear Probing} & \multicolumn{3}{c}{Fine-tuning} \\
 & AP$_{50}$ & AR$_{50}$ & AUC & AP$_{50}$ & AR$_{50}$ & AUC \\ \shline
AttMask~\cite{kakogeorgiou2022hide_attmask} & 26.6 & 60.1 & 0.593 & 66.6 & 26.3 & 0.785 \\
MST~\cite{li2021mst} & 26.2 & 57.1 & 0.575 & 28.2 & 64.0 & 0.732 \\
iBot~\cite{zhou2022image} & 23.0 & 65.0 & 0.620 & 46.7 & 71.8 & 0.819 \\
SMIT~\cite{jiang2022self_SMIT} & 21.8 & 64.0 & 0.629 & 72.7 & 65.5 & 0.859 \\
SMART & 28.8 & 65.5 & 0.642 & 66.8 & 79.5 & 0.895 \\ \hline
\end{tabular}%
}
\end{table}

\begin{table}[t]
\caption{Immunotherapy (132/68)$^*$}
\label{tab:results_classification_immunotherapy}
\def\arraystretch{1.25}
\resizebox{0.48\textwidth}{!}{%
\begin{tabular}{lllllll}
\multirow{2}{*}{\begin{tabular}[c]{@{}l@{}}Pretext SSL\\ Task\end{tabular}} & \multicolumn{3}{c}{Linear Probing} & \multicolumn{3}{c}{Fine-tuning} \\
 & AP$_{50}$ & AR$_{50}$ & AUC & AP$_{50}$ & AR$_{50}$ & AUC \\ \shline
AttMask~\cite{kakogeorgiou2022hide_attmask} & 44.9 & 54.0 & 0.570 & 56.0 & 68.5 & 0.660 \\
MST~\cite{li2021mst} & 42.5 & 52.0 & 0.550 & 48.6 & 59.1 & 0.630 \\
iBot~\cite{zhou2022image} & 45.9 & 57.5 & 0.580 & 55.6 & 65.5 & 0.630 \\
SMIT~\cite{jiang2022self_SMIT} & 46.4 & 58.1 & 0.620 & 56.3 & 67.0 & 0.660 \\
SMART & 54.5 & 68.4 & 0.660 & 57.4 & 71.7 & 0.740 \\ \hline
\end{tabular}%
}
\end{table}


\subsection{SMART on limited data regime}
\begin{figure}[t]
    \centering
    \includegraphics[width=0.95\columnwidth]{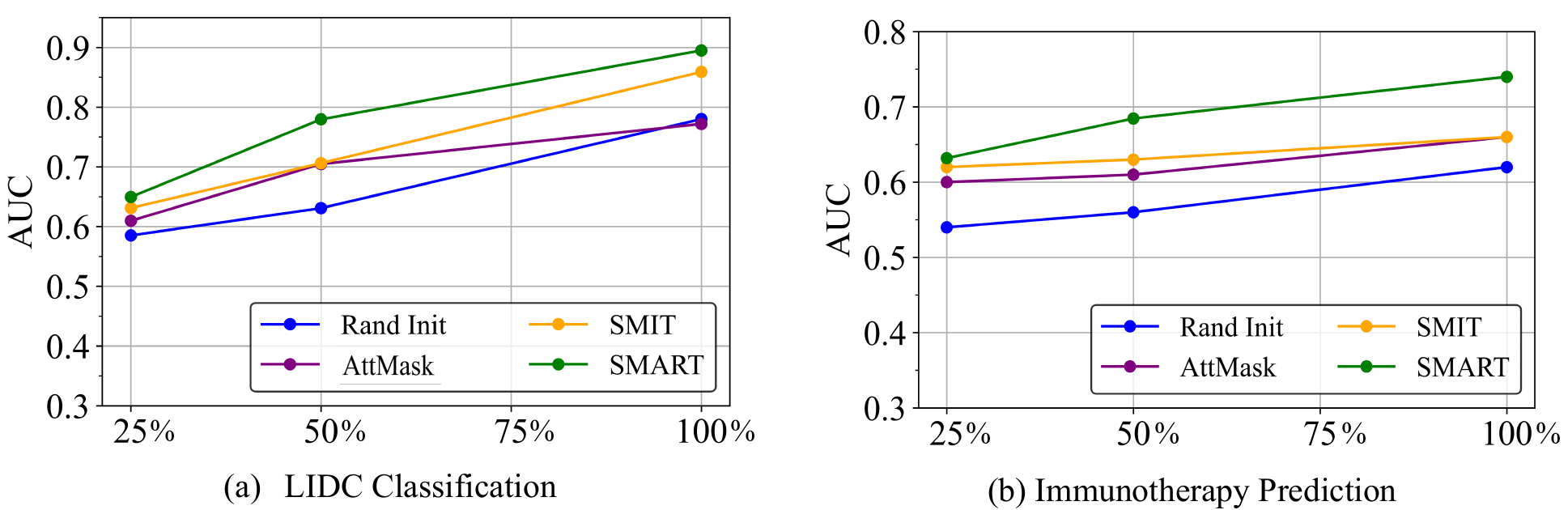}
    \caption{Data limited training accuracy.  }
    \label{fig:fewshotclassification}
\end{figure}
Performance on learning under limited data was evaluated using randomly sampled examples from the respective training sets at percentile of 25\% and 50\% and then evaluating on the testing sets (Figure \ref{fig:fewshotclassification}) for LIDC classification and Immunotherapy prediction tasks. Notably, SMART consistently outperformed all other methods on both tasks, regardless of the number of examples used. Also, it was more accurate than AttMask even with only 50\% of the examples for both classification and segmentation tasks. SMIT was less accurate than SMART, but was more accurate than AttMask at the 50\% example scenario for the segmentation but not the classification task. In addition, SMART's accuracy improved at a faster rate with increasing number of samples. SMIT was the second most accurate method, but showed slower rate of accuracy improvement than SMART.

\begin{figure}[t]
    \centering
    \includegraphics[width=\columnwidth]{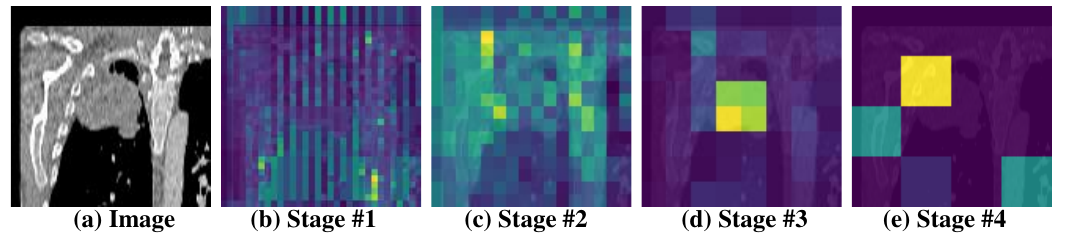}
    \caption{Attention map for SMART when the semantic-attention module was put at different stages of Swin.}
    \label{fig:att_diff_stages}
\end{figure}

\section{Ablation Studies}
We effectively analyze the sub-parts of SMART, its advantages and discuss the limitations in this section.

\subsection{Impact of individual losses}
\label{sup_subsec:individualosses}

We analyzed the impact of various losses on fine tuning accuracy on LIDC data set, shown in Table~\ref{tab:howdodifferentlossesfare1} and~\ref{tab:howdodifferentlossesfare}. We observed two key things: (1) AMIP was crucial for the network to perform well and achieve high accuracy, (2) including GITD (and in turn, Noisy Teacher) resulted in the next best model post the semantic attention model, indicating GITD importance for learning a good set of pretraining weights.

\subsection{Impact of semantic attention module}
\begin{figure}[t]
    \centering
    \includegraphics[width=0.9\columnwidth]{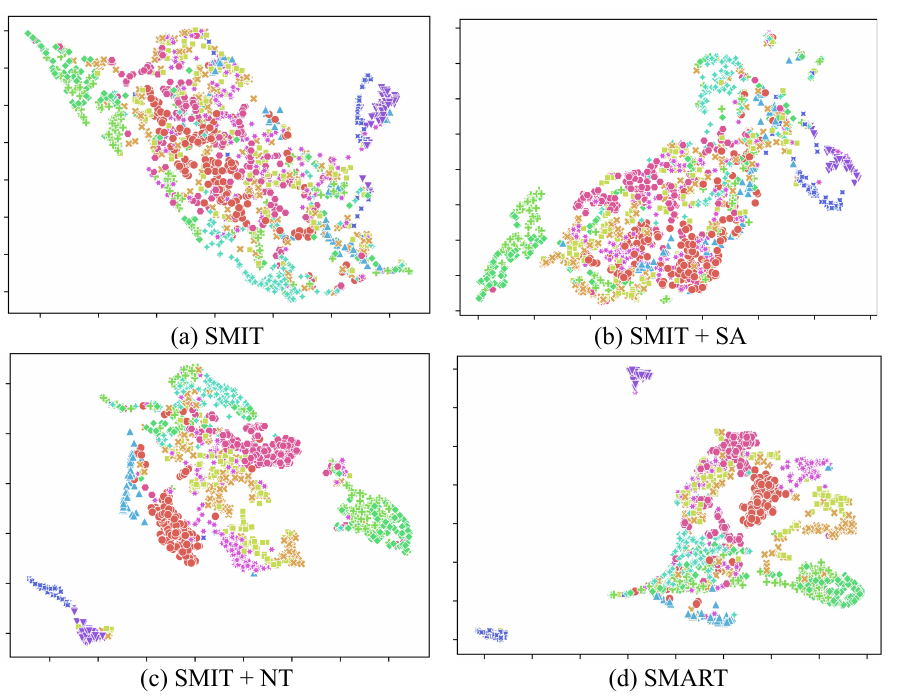}
    \caption{uMAP clusters computed from different steps towards SMART and applied to organMNIST3D. NT: noisy teacher, SA: semantic attention.}
    \label{fig:umap}
\end{figure}

We investigate the influence of semantic attention on seperating organs using organminst3D data set, shown in Fig.~\ref{fig:umap}. From Fig.~\ref{fig:umap}a and Fig.~\ref{fig:umap}b, adding semantic attention improved the separation of organs more than with SMIT alone. SMART using only SA resulted in a lower intra-cluster distance of 1.43 $\pm$ 0.49 that was lower than SMIT of 1.53 $\pm$ 0.45 and a higher inter-cluster distance of 4.18 $\pm$ 2.74, shown in Table \ref{tab:smart_othertasks_stages}. These results indicate that sematic class attention improves accuracy more than when just using standard Swin configuration.

\subsection{Impact of semantic attention module location}
\begin{figure}[t]
    \centering
    \includegraphics[width=1.0\columnwidth]{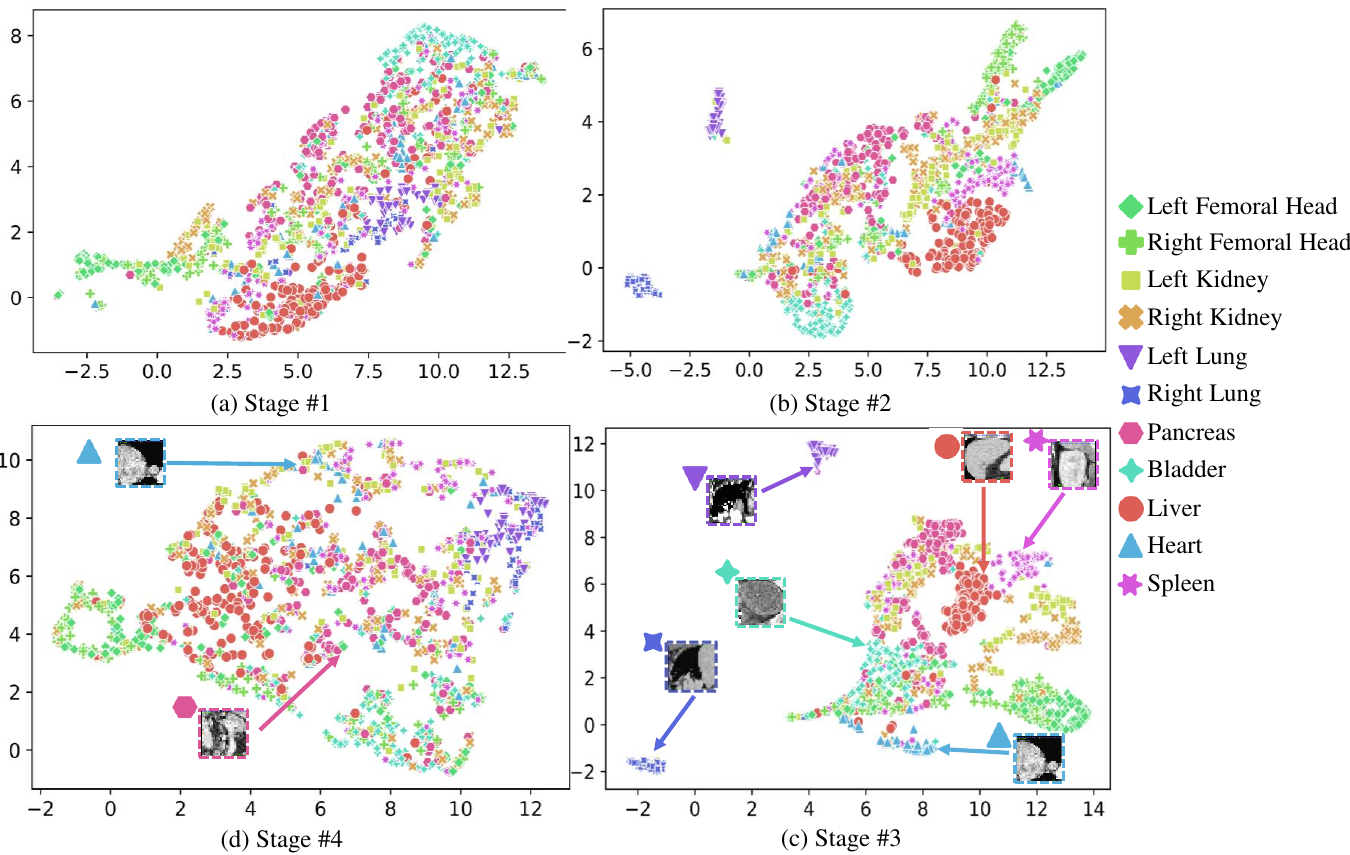}
    \caption{SA Location on the organMinist3D data}
    \label{fig:smart_SAlocation}
\end{figure}

We investigated the impact of semantic attention module location on the learnt representation using the miniorgan3D data, shown in figure \ref{fig:smart_SAlocation}. The inter-and intra cluster distance are shown in Table \ref{tab:smart_othertasks}. Putting the semantic attention module at Swin Stage of 3 lead to the largest inter-cluster and minimum intra-cluster distance. We also visualized the attention map, when the semantic attention modeule was placed at different stages of Swin, in Figure \ref{fig:att_diff_stages}. Putting the attention module at stage 3 leands to the most semantic attentions than putting it at other stages.

\begin{figure}[t]
    \centering
    \includegraphics[width=0.5\textwidth]{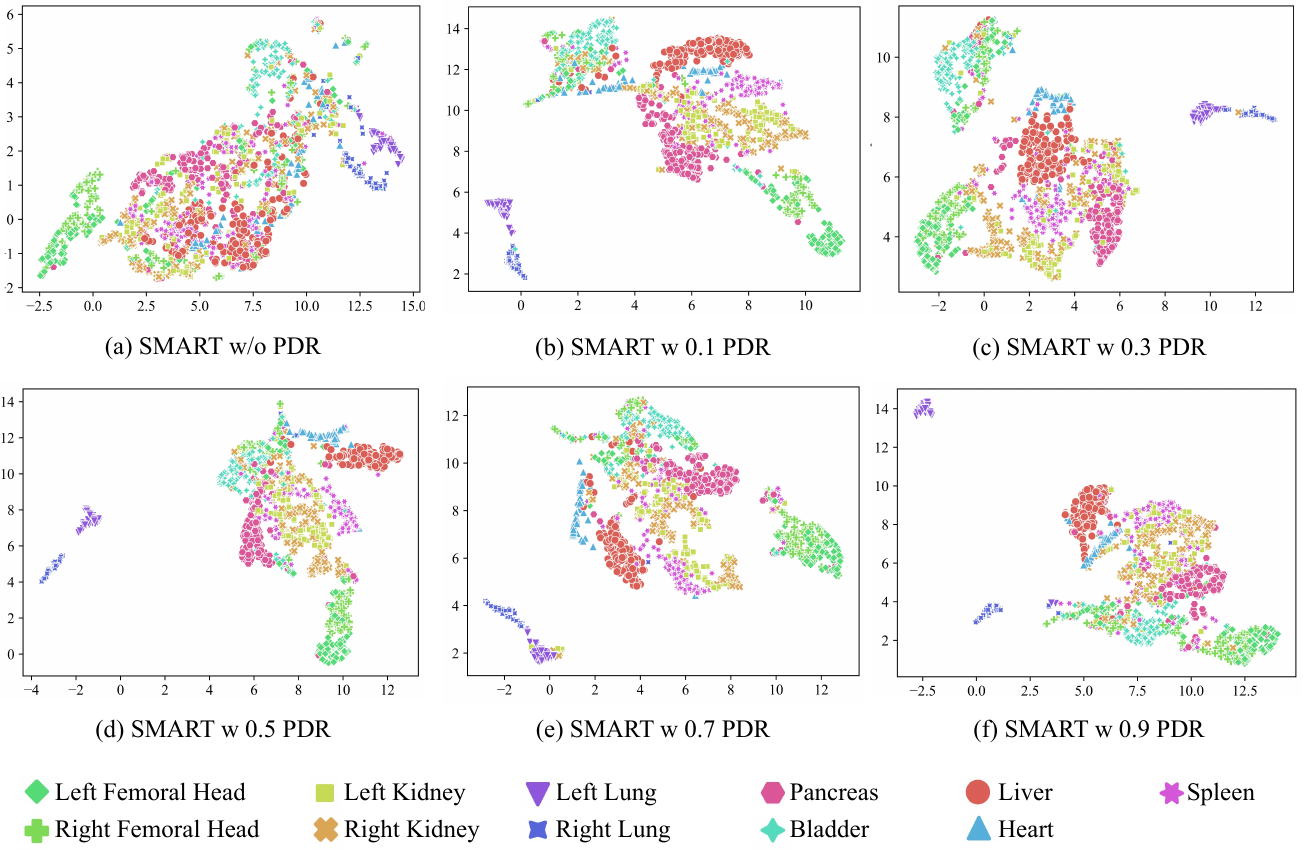} 
    \caption{Unsupervised clustering results produced using SMARTFormer pretrained with different amounts of patch drop ratios (PDR).}
    \label{fig:att_msk_comp_patchdropratios}
\end{figure}

\begin{table}[t]
        \caption{Impact of losses, semantic attention, and noisy teacher on Task 1 with the LIDC dataset.}
    \begin{subtable}[t]{0.42\columnwidth}
        \def\arraystretch{1.25}
        \scriptsize
        \caption{Key developments}
        \label{tab:howdodifferentlossesfare1}
        \resizebox{\columnwidth}{!}{%
        \begin{tabular}{lll}
        \begin{tabular}[c]{@{}l@{}}Noisy\\ Teacher\end{tabular} & \begin{tabular}[c]{@{}l@{}}Semantic \\ Attention\end{tabular} & AUC \\ \shline
        $\times$ & $\times$ & 0.859 \\
        $\checkmark$ & $\times$  & 0.882 \\
        $\times$ & $\checkmark$ & 0.875 \\
        $\checkmark$ & $\checkmark$ & 0.895 \\ \shline
        \end{tabular}%
        }        
     \end{subtable}
    \hfill
    \begin{subtable}[t]{0.55\columnwidth}
        \def\arraystretch{1.25}
        \scriptsize
        \caption{The role of different losses}
        \label{tab:howdodifferentlossesfare}
        \resizebox{\columnwidth}{!}{%
        \begin{tabular}{lllll}
        AMIP & AMPD & AITD & GITD & AUC \\ \shline
        $\checkmark$ & $\times$ & $\times$ & $\times$ & 0.859\\
        $\checkmark$ & $\checkmark$ & $\times$ & $\times$ & 0.869 \\
        $\checkmark$ & $\checkmark$ & $\checkmark$ & $\times$ & 0.878 \\
        $\checkmark$ & $\times$ & $\checkmark$ & $\checkmark$ & 0.865 \\
        $\times$ & $\checkmark$ & $\checkmark$ & $\checkmark$ & 0.829 \\
        $\checkmark$ & $\checkmark$ & $\checkmark$ & $\checkmark$ & 0.895 \\ \shline
        \end{tabular}%
        }
    \end{subtable}

\end{table}

\subsection{Impact of Noisy Teacher}

\textbf{Noisy teacher produced more diverse pretrained features.} We further plot the UMAP clusterings for OrganMNIST3D. SMART without both its component can be considered as its predecessor, SMIT, and we use the image tokens produced after average pooling of the features from stage \#4 layers for representations. As shown in Fig.~\ref{fig:umap}b, addition of noisy teacher regularization for SMART pretraining was better at separating the organs better than just SMIT, (Fig.~\ref{fig:umap}a) as well as when the network was pretrained without noisy teacher (Fig.~\ref{fig:umap}c). Inclusion of noisy teacher reduced the intra-cluster distance to 1.46 $\pm$ 0.30 and increased inter-cluster distance to 5.17 $\pm$ 2.98, indicating that diversifying tokens during self-distillation is important to improve the ability of the network to distinguish salient image regions. 

\textbf{Noisy teacher produced more diverse Attention maps.}  Fig. \ref{fig:semanticattn_eval_main} shows the pretrained and fine-tuned semantic attention of SMARTFormer and SMARTFormer w/o noise. In the pretrained semantic attention map, adding PDR noise to teacher network, SMARTFormer learned to attend towards tumor regions far more effectively than when trained without adding noise. This is also reflected in the localization of tumors both in pretraining and following fine-tuning of SMARTFormer network for predicting immunotherapy response. As shown, addition of noise improves the localization ability compared to the same network trained without noise, which results in focus on irrelevant normal tissue locations. AttMask results are also shown for comparison, which shows a preferential localization within the lung instead of the tumors. Training of the network from scratch results in a non-preferential localization towards different regions in the image.

\begin{figure}[t]
    \centering
    \includegraphics[width=0.75\columnwidth]{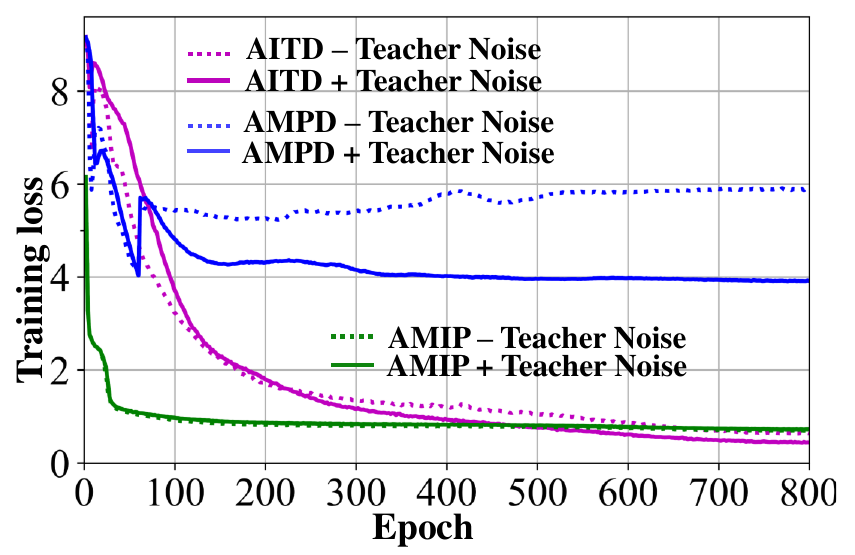}
    \caption{Pre-Training curve with adding noise to the teacher network.}
    \label{fig:pretrainig_curve}
\end{figure}

\textbf{Noisy teacher produced more stable pretraining curve.}  We also illustrated the pre-training curve when noise was added to the teacher in Figure \ref{fig:pretrainig_curve}. Introducing teacher noise to the network resulted in more stable AMPD training, as shown by the comparison between the dotted blue line and the solid blue line in Figure \ref{fig:pretrainig_curve}.

\subsection{Impact of Patch Dropout Ratio (PDR) for Noisy Teacher}
\label{sup_subsec:noisyTeacher}
Fig. \ref{fig:att_msk_comp_patchdropratios} shows the UMAP projection of SMARTFormer with 0.0, 0.1, 0.3, 0.5, 0.7, 0.9 of PDR. With PDR of 0.0, the projected features of different organs collapse together while adding noise leading a better cluster separation. We observed that small amounts of noise or patch drop ratio or PDR $<$ 0.5 had minimal impact on improving the separation of organs, PDR of 0.7 resulted in the best separation, while further increasing PDR led to poor separation with unsupervised clustering. With PDR of 0.0, the projected features of different organs collapse together while adding noise leading a better cluster separation.

\section{Insightful Findings}
\label{sup_subsec:micdrop}

\begin{table}[t]
\def\arraystretch{1.2}
\scriptsize
\centering
\caption{Impact of semantic attention module location}
\label{tab:smart_othertasks_stages}
\resizebox{\columnwidth}{!}{%
\begin{tabular}{lll||ll}
SSL Task  & Intra-cluster ($\downarrow$) & Inter-cluster ($\uparrow$) &LIDC Prob &LIDC FT\\ \shline
\hline
\hspace{0.5em}SMART S1 & 1.83 $\pm$ 0.41 & 2.78 $\pm$ 1.56 &0.581&0.597\\
\hspace{0.5em}SMART S2 & 1.66 $\pm$ 0.58 & 5.21 $\pm$ 3.75 &0.615&0.647\\
\hspace{0.5em}SMART S3 (Default)  & 1.28 $\pm$ 0.57 & 5.93 $\pm$ 3.30 & \textbf{0.642}& \textbf{0.895}\\ 
\hspace{0.5em}SMART S4  & 1.41 $\pm$ 0.30 & 4.12 $\pm$ 2.51 &0.605&0.684\\
\hline
\end{tabular}%
}

\end{table}


\textbf{[CLS] token versus Averaging pooling for SWIN:}\\
We studied whether the [CLS] tokens derived from SA block benefit downstream classification accuracy more than average pooled tokens derived from stage \#4. We fine tuned the networks for LIDC tumor classification with different configurations. Our analysis showed that model using pretrained [CLS] tokens was the most accurate (AUC of 0.895, AP50 of 0.668 and AR50 of 0.795). The network trained using tokens extracted from stage \#4 on the other hand was the least accurate (AUC of 0.780, AP50 of 0.632 and AR50 of 0.436) and less accurate than the network using randomly initialized SA block (AUC of 0.875, AP50 of 0.621 and AR50 of 0.700), which clearly shows the benefit of semantic attention.

\textbf{Does using explicit lung disease CT scans during pretraining affect down-stream performance?}\\
Specifically, we examined the effect of using solely lung disease-based CT scans (7,289 samples), versus using CT scans derived from non-lung based diseases (3,123 samples) for pretraining, to downstream classification performance. Our findings showed notable improvements in terms of area under the receiver operating characteristic curve denoted for short as AUC. When exclusively using lung disease CTs, SMART achieved AUC of 0.883 for LIDC tumor classification. Utilizing only non-lung disease CTs yielded a slightly lower AUC of 0.861. On the other hand, both of these models were more accurate than the model that was trained from scratch with an AUC of 0.780.

\textbf{Does SMART generalize to other network architectures and pretext self-supervised learning tasks ? }

We investigated whether SMART could be applied to other pretext SSL tasks and Vision Transformers (ViTs) besides Swin, as shown in Table \ref{tab:smart_othertasks}. On the iBot task, incorporating SMART with iBot resulted in an increase in fine-tuning accuracy from 0.819 to 0.881 and a linear probing accuracy improvement from 0.620 to 0.xx. For the ViT architecture, SMART achieved the highest linear probing accuracy of 0.662 compared to SMIT's 0.643. Overall, SMART demonstrated accuracy improvements in both ViT and Swin architectures for linear probing and fine-tuning settings.

\begin{table}[t]
\def\arraystretch{1.2}
\scriptsize
\centering
\caption{SMART generalized to other pretext SSL tasks and network structures}
\label{tab:smart_othertasks}
\resizebox{\columnwidth}{!}{%
\begin{tabular}{llll||ll}
SSL Task & Network & Intra-cluster ($\downarrow$) & Inter-cluster ($\uparrow$) &LIDC Prob &LIDC FT\\ \shline
iBot~\cite{zhou2022image} & Swin & {2.13 $\pm$ 0.37} & {3.29 $\pm$ 2.07}&0.620 &0.819\\
iBot+SMART & Swin&1.06$\pm$0.34& 5.36$\pm$3.16 & \textbf{0.656}&0.881\\ 
\hline
\hline
SMIT~\cite{jiang2022self_SMIT} & ViT & 1.30$\pm$0.51& 6.44$\pm$5.27 &0.643 &0.816\\
SMART & ViT&1.15$\pm$0.41  & 6.59$\pm$4.60&\textbf{0.662} &0.847\\
SMIT~\cite{jiang2022self_SMIT} & Swin& {1.53 $\pm$ 0.45} & {3.73 $\pm$ 2.74}&0.629 &0.859\\
SMART & Swin & 1.28 $\pm$ 0.57 & 5.93 $\pm$ 3.30& \textbf{0.642}& \textbf{0.895}\\ \hline
\end{tabular}%
}
\end{table}

\section{Discussion and conclusion}
We introduced an attention-guided masking approach for Swin transformers that allows selective masking during pretraining. Our approach also allows to visualize attention maps for downstream task, thus increasing the interpretability of Swin transformers. Interpretability of deep learning model results are crucial for clinician acceptance and eventual deployment of medical applications. We found that noisy teacher in combination with semantic attention was highly effective in improving pretraining and downstream accuracy. We also found that masked image prediction was crucial for pretraining in comparison to self-distillation losses. Our results also show that our approach can achieve reasonably high accuracy with fewer labeled examples in downstream tasks and improves over ViT-based selective masking methods.
\\
Our present limitations lie in fully understanding the learning dynamics between the student and teacher models and effectively leveraging the gained knowledge for improved self-supervised and semi-supervised learning, particularly in medical settings where data is naturally scarce. Utilizing the learnt knowledge for research in data and compute efficient models is future work.

{
\small
\bibliographystyle{IEEEtran}
\bibliography{main,mybibliography}
}

\end{document}